\documentclass[10pt,journal,compsoc,onecolumn]{IEEEtran}

\usepackage{graphicx}
\usepackage{amsmath,amssymb,amsfonts,bm}
\usepackage{amsthm}
\usepackage{mathrsfs}
\usepackage{xcolor}
\usepackage{textcomp}
\usepackage{booktabs}
\usepackage{algorithm}
\usepackage{algorithmicx}
\usepackage{algpseudocode}
\usepackage{listings}
\usepackage{enumitem}
\usepackage{upgreek}
\usepackage{longtable}

\theoremstyle{plain}

\begin{document}
	
	\title{An active-learning framework for real-time depth perception from monocular vision streams}
	
	\author{Xiaorong~Zeng, Weiqiang~Chen, Peng~Shi,~\IEEEmembership{Fellow,~IEEE,} Liang~Su, \\ Zirui~Wang, Xuewu~Ji, and~Shuiwen~Shen% <-this % stops a space
		\IEEEcompsocitemizethanks{
			\IEEEcompsocthanksitem X. Zeng, Z. Wang, and S. Shen are with the School of Mechanical and Automotive Engineering, Xiamen University of Technology, Xiamen, Fujian 361024, China. S. Shen is also with the Xiamen Innovative Centre for Automotive Electric Driving, Xiamen 361010, China (Corresponding author: Shuiwen Shen, e-mail: 2023000085@xmut.edu.cn).
			\IEEEcompsocthanksitem W. Chen and L. Su are with King Long United Automotive Industry Co. Ltd., Xiamen 361023, China.
			\IEEEcompsocthanksitem P. Shi is with the School of Electrical and Electronic Engineering, The University of Adelaide, Adelaide, SA 5005, Australia.
			\IEEEcompsocthanksitem X. Ji is with the School of Vehicle and Mobility, Tsinghua University, Beijing 100084, China.
		}%
	}
	
	\IEEEtitleabstractindextext{%
		\begin{abstract}
			Biological visual systems can perceive depth from monocular vision flow, continuously integrating temporal visual cues while maintaining a balance between stability and plasticity in dynamic environments. In contrast, artificial perception models deployed on resource-constrained edge devices are typically trained in a static offline manner and remain frozen after deployment, often suffering severe performance degradation under domain shifts. While large-scale models may encode broad knowledge through massive parameter redundancy, lightweight networks face a static optimization dilemma: forcing compact models to learn universal geometric representations is computationally inefficient and often leads to performance saturation. To resolve this issue, an Online Active Learning (\textbf{OAL}) mechanism is introduced to endow compact neural networks with the capability to adapt continuously during operation. A closed-loop \textbf{Predict--Evaluate--Correct} learning paradigm is established to actively select high-confidence, information-rich signals from streaming visual input. Crucially, Elastic Weight Consolidation (\textbf{EWC}) is employed not merely to prevent catastrophic forgetting, but to enforce \textbf{Selective Plasticity}, preserving parameters that encode globally relevant structural knowledge while allowing local alignment to newly observed environments. Built upon a MobileNetV3-Small backbone, the proposed system achieves approximately a $75\%$ reduction in computational cost while maintaining competitive depth estimation accuracy. Experimental results demonstrate that adaptability is not solely determined by model size, but rather by how effectively parameter plasticity is regulated in dynamic environments.
		\end{abstract}
		
		}
	
	\maketitle
	
	\IEEEdisplaynontitleabstractindextext
	\IEEEpeerreviewmaketitle
	
	% ==========================================
	% SECTION I: INTRODUCTION
	% ==========================================
	\section{Introduction}
	\label{sec:introduction}
	\IEEEPARstart{M}{onocular} depth estimation is a fundamental problem in 3D computer vision and an essential component of spatial perception for autonomous vehicles, intelligent robots, and edge artificial intelligence systems. Recent advances in self-supervised learning have substantially improved monocular depth prediction by removing the need for expensive LiDAR-based ground-truth annotations. Nevertheless, most existing methods still follow a static ``train-then-deploy'' paradigm, in which model parameters remain fixed after deployment. This limitation becomes particularly critical in real-world driving scenarios, where illumination changes, weather variations, and cross-domain shifts can significantly degrade performance. Therefore, moving beyond static optimization toward continuous post-deployment adaptation is a necessary step for robust real-world monocular perception.
	
	\subsection{Perceptual Plasticity in Biological and Artificial Systems}
	Biological vision provides a useful conceptual analogy for adaptive perception systems. In particular, predictive coding suggests that perception can be understood as a continuous process of prediction, error evaluation, and internal correction \cite{friston2005theory, rao1999predictive}. This closed-loop mechanism allows biological systems to remain responsive to dynamic environments while preserving stable perceptual functions \cite{mermillod2013stability}. By contrast, most deep neural networks for monocular depth estimation are optimized offline and then kept fixed during deployment. Although large over-parameterized models can partially absorb domain variation through extensive offline training, this strategy is often impractical for lightweight edge-oriented systems operating under strict computational and parameter constraints.
	
	\subsection{The Static Optimization Dilemma in Lightweight Models}
	A key challenge in deploying lightweight monocular depth estimation models is that their limited representational capacity makes static offline optimization inherently restrictive. Unlike large-scale models with substantial parameter redundancy, compact networks operate under tight computational and parameter budgets. As a result, requiring such models to absorb broad environmental variation entirely during offline training is often computationally inefficient and may lead to underfitting or early performance saturation. This limitation is referred to here as the \textbf{Static Optimization Dilemma}. Empirical observations further show that commonly adopted enhancement strategies, such as aggressive data augmentation and additional attention modules, do not consistently improve lightweight models and may even degrade performance by introducing optimization conflicts. These observations suggest that robust deployment on edge platforms cannot rely solely on stronger offline training, but instead requires an adaptive mechanism that continues to correct the model after deployment.
	
	\subsection{Resolving the Dilemma via Gated Structure and Selective Plasticity}
	To overcome the above limitation, a unified framework is developed to combine structural efficiency with continuous online adaptation. First, a Gated Cross-scale Additive Fusion (\textbf{GCAF}) module is introduced to replace the conventional skip concatenation in the U-Net-style depth decoder \cite{ronneberger2015u}. For lightweight backbones such as \textbf{MobileNet}V3-Small \cite{howard2019searching}, this design reduces redundant feature transmission and suppresses high-frequency noise from shallow encoder layers, thereby improving representation efficiency under strict computational constraints.
	
	Second, the framework adopts an online adaptation strategy based on a closed-loop \textbf{Predict--Evaluate--Correct} mechanism. Within this loop, scene geometry is predicted from incoming monocular frames, the reliability of the prediction is evaluated through photometric consistency, and incremental correction is then performed using high-confidence observations from the video stream. To stabilize this process, \textbf{Elastic Weight Consolidation} (\textbf{EWC}) \cite{kirkpatrick2017overcoming} is introduced as a selective plasticity constraint. Parameters located in the \textbf{high-Fisher information space}, which encode previously learned global structural knowledge, are strongly constrained to preserve learning stability, whereas parameters in the low-Fisher subspace remain more flexible for local domain alignment. In this way, the proposed framework improves both the efficiency and the adaptability of lightweight monocular depth estimation.
	
	Furthermore, the stable geometric representations learned through this CTTA paradigm establish a reliable foundation for downstream spatial perception tasks, such as pseudo-LiDAR-based 3D object recognition.
	
	\subsection{Contributions}
	The main contributions of this paper are summarized as follows:
	\begin{itemize}
		\item \textbf{A lightweight and efficient depth estimation architecture.} A Gated Cross-scale Additive Fusion (\textbf{GCAF}) module is introduced to replace conventional dense skip concatenation with a selective semantics-guided fusion strategy. By explicitly regulating cross-scale information flow, this design improves feature aggregation efficiency, reduces the sensitivity of lightweight backbones to high-frequency noise, and provides a more effective representation for resource-constrained monocular depth estimation.
		
		\item \textbf{An Online Active Learning (\textbf{OAL}) framework based on a closed-loop \textbf{Predict--Evaluate--Correct} mechanism.} The proposed OAL framework integrates depth prediction, confidence evaluation, and incremental parameter adaptation within a unified self-supervised loop. It enables continuous online correction from streaming monocular video without requiring external supervision or offline domain-specific retraining.
		
		\item \textbf{A selective plasticity mechanism for stable online adaptation.} Elastic Weight Consolidation (\textbf{EWC}) is incorporated not merely to reduce catastrophic forgetting, but to preserve parameters in the \textbf{high-Fisher information space}, which encode important global structural priors, while allowing more flexible adaptation in the low-Fisher subspace. This mechanism provides a principled way to balance adaptation and stability during online deployment.
	\end{itemize}

	% ==========================================
	% SECTION II: RELATED WORK
	% ==========================================
	\section{Related Work}
	\label{sec:related_work}
	Monocular depth estimation has progressed from early supervised methods \cite{eigen2014depth, fu2018deep} to self-supervised learning frameworks that remove the need for dense depth annotations \cite{garg2016unsupervised, zhou2017unsupervised, godard2017unsupervised, yin2018geonet}. More recent studies have further improved performance through increasingly sophisticated architectures, including 3D packing strategies \cite{guizilini20203d} and self-attention-based designs \cite{johnston2020self}. To improve deployment efficiency, a number of lightweight depth estimation models have also been proposed for resource-constrained platforms \cite{wofk2019fastdepth, bhat2021adabins, ranftl2020towards, zhang2023lite}. Although these compact models achieve favorable static efficiency, their performance often degrades when the deployed environment differs from the offline training distribution.
	
	To address domain shift, prior studies have explored test-time adaptation (TTA) \cite{sun2020test, wang2020tent, liang2020we}, online meta-learning \cite{li2020online}, and unsupervised domain adaptation methods for depth perception \cite{casser2019depth, kundu2018adadepth, zheng2018t2net, tonioni2019learning}. However, straightforward online fine-tuning on streaming visual data is often unstable and can lead to catastrophic forgetting \cite{li2017learning}. More general continual learning methods, including Synaptic Intelligence \cite{zenke2017continual}, Gradient Episodic Memory \cite{lopezpaz2017gradient}, iCaRL \cite{rebuffi2017icarl}, and replay-based strategies \cite{delange2021continual, parisi2019continual}, provide useful mechanisms for preserving prior knowledge, but they are rarely designed for continuous geometric regression problems such as monocular depth estimation.
	
	More recently, continual test-time adaptation (CTTA) has been developed to handle continuously changing target distributions. Representative methods include stochastic weight restoration (CoTTA \cite{wang2022continual}), memory-efficient meta-networks (EcoTTA \cite{song2023ecotta}), robust statistics estimation (RoTTA \cite{yuan2023robust}), and anti-forgetting regularization strategies such as EATA \cite{niu2022efficient} and SAR \cite{niu2023towards}. Although these approaches have shown strong performance in image classification and semantic segmentation, many of them depend on entropy-based objectives, teacher--student pseudo-labeling, or repeated test-time augmentation. Such designs do not transfer naturally to continuous depth regression and may also introduce substantial computational overhead during deployment. Therefore, achieving stable and efficient online adaptation for lightweight monocular depth estimation remains an open problem.

	% ==========================================
	% SECTION III: PROPOSED FRAMEWORK
	% ==========================================
	\section{Proposed Framework}
	\label{sec:methods}
	
	\subsection{Overall Framework}
\begin{figure}[!htbp]
		\centering
		\includegraphics[width=0.95\linewidth]{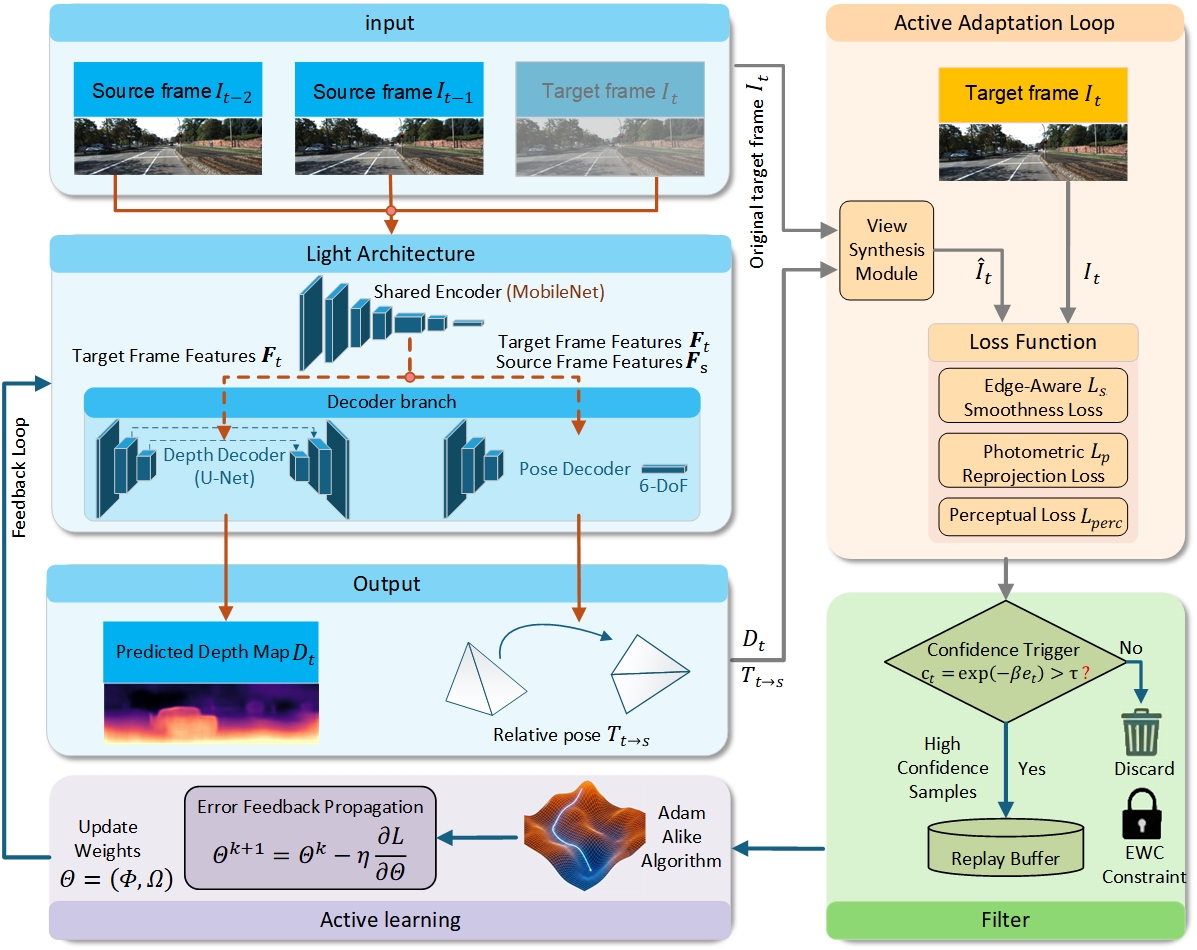}
		\caption{Overall architecture of the proposed active learning monocular depth estimation framework.}
		\label{fig:arch_overview}
	\end{figure}

	The proposed Online Active Learning (\textbf{OAL}) framework is built upon a closed-loop \textbf{Predict--Evaluate--Correct} mechanism for self-supervised monocular depth adaptation. As illustrated in Fig.~\ref{fig:arch_overview}, the framework updates the deployed model by minimizing the photometric reprojection error between the observed target frame and the reconstructed target frame generated from adjacent source frames. In this way, the model can continuously refine its depth prediction during deployment without requiring ground-truth depth supervision.
	
	To support real-time operation on resource-constrained platforms, the framework combines a lightweight backbone with an \textbf{AdamW}-based \cite{loshchilov2018decoupled} online optimization module for incremental parameter updates. Rather than relying solely on exhaustive offline optimization, the proposed framework enables the deployed model to remain responsive to changing visual conditions through continuous self-correction from streaming monocular video.
	
	Specifically, the proposed \textbf{OAL} framework consists of four main components: 1) a depth estimation network, 2) a pose estimation network, 3) a loss computation module, and 4) an online optimization module. The depth network predicts dense depth maps from sequential monocular images, while the pose network estimates the relative six-degree-of-freedom (6-DoF) camera motion between adjacent frames. The loss computation module evaluates photometric and geometric consistency, and the online optimization module updates the model parameters according to the resulting self-supervised error signal.
	
	\subsection{Principle of OAL-based Monocular Depth Estimation}
	\begin{sloppypar}
	The proposed Online Active Learning (\textbf{OAL}) framework applies the \textbf{Predict--Evaluate--Correct} mechanism to self-supervised monocular depth estimation through differentiable view synthesis. Specifically, scene geometry and camera motion are predicted from adjacent monocular frames, the prediction is evaluated by reconstructing the target frame from neighboring source frames, and the resulting photometric reprojection error is then used as a feedback signal for online correction. In this way, OAL enables continuous adaptation during deployment without requiring ground-truth depth supervision.
	\end{sloppypar}
	
	The key idea is that, if the predicted depth and relative pose are accurate, the reconstructed target frame should be photometrically consistent with the observed target frame. Therefore, the reprojection error provides a self-supervised measure of prediction quality and serves as the basis for incremental parameter updates in the online learning loop.
	
	\subsubsection{Real-time Depth Learning Paradigm}
	\label{OALParadigm}
	Consider two consecutive frames in a monocular video stream. Let $I_t$ denote the target frame at time step $t$, and let $I_s$ denote an adjacent source frame, where $s=t-1$ in the causal online setting. The objective is to reconstruct $I_t$ from $I_s$ using the predicted scene geometry and relative camera motion. To this end, the framework includes two subnetworks:
	\begin{itemize}
		\item a depth estimation network $\textrm{fNN}_{dp}(\Phi)$, which predicts the depth maps $D_t$ and $D_s$; and
		\item a pose estimation network $\textrm{fNN}_{tf}(\Omega)$, which predicts the relative camera transformation $T_{t \to s}$.
	\end{itemize}
	
	The relative transformation $T_{t \to s}$ is modeled as a rigid-body motion consisting of a rotation matrix $R$ and a translation vector $t$:
	\begin{equation}
		T_{t \to s} =
		\begin{bmatrix}
			R & t \\
			0 & 1
		\end{bmatrix}
	\end{equation}
	
	To reconstruct the target view, a differentiable geometric projection pipeline is adopted.
	
\begin{figure}[htbp]
		\centering
		\includegraphics[width=0.8\linewidth]{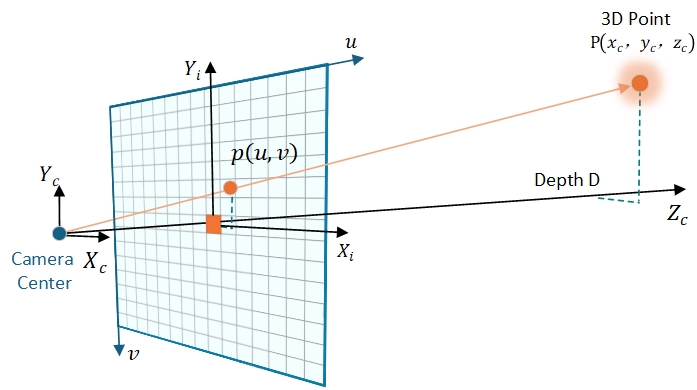}
		\caption{Back-projection process.}
		\label{fig:back_proj}
	\end{figure}

	\textbf{Step 1: Back-projection.} Given the camera intrinsic matrix $K$, a homogeneous pixel coordinate $p_{t-1} = [u, v, 1]^T$ in the source frame $I_{t-1}$ is back-projected into 3D space using the predicted depth value $D_{t-1}(u,v)$. The corresponding 3D point $P_{t-1}$ in the source camera coordinate system is computed as
	\begin{equation}
		P_{t-1} = D_{t-1}(u,v) K^{-1} p_{t-1}
	\end{equation}
	which recovers the local scene structure from the 2D image plane, as illustrated in Fig.~\ref{fig:back_proj}.
	
	\textbf{Step 2: Coordinate Transformation.} The reconstructed 3D point $P_{t-1}$ is then transformed from the source camera coordinate system to the target camera coordinate system through the estimated relative pose $T_{t \to s}$:
	\begin{equation}
		P_t = T_{t \to s} P_{t-1} = R P_{t-1} + t
	\end{equation}
	
	\textbf{Step 3: Re-projection.} After transformation, the 3D point $P_t$ is projected onto the image plane of the target frame to obtain the corresponding continuous image coordinate $\hat{p}_t$:
	\begin{equation}
		\hat{p}_t \simeq D_t^{-1} K P_t
	\end{equation}
	By combining the above steps, the geometric mapping from the source pixel location to the target image coordinate can be written as
	\begin{equation}
		\hat{p}_t \sim D_t^{-1} K T_{t \to s} D_{t-1} K^{-1} p_{t-1}
		\label{eq:projection_geometry}
	\end{equation}
	
\begin{figure}[!htbp]
		\centering
		\includegraphics[width=0.8\linewidth]{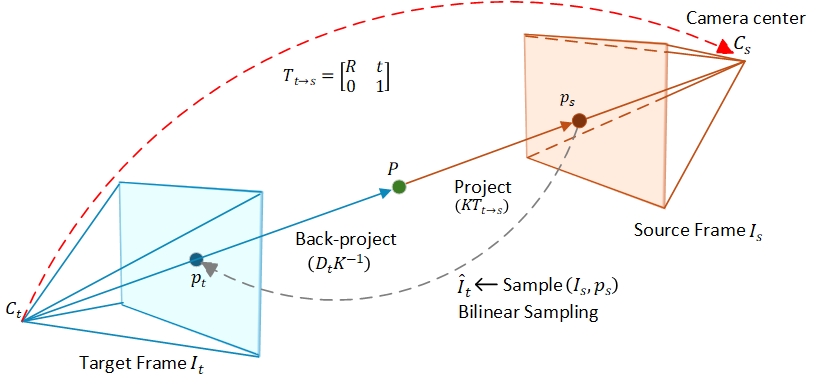}
		\caption{\textbf{Schematic of View Synthesis.} Illustration of the full photometric consistency loop described in Eqn.~(\ref{eq:projection_geometry}). A pixel from the target frame $I_t$ is back-projected, transformed to the source frame $I_s$ via pose $T_{t\to s}$, and re-projected to sample the color value.}
		\label{fig:view_synthesis}
	\end{figure}

	\textbf{Step 4: View Synthesis via Bilinear Sampling.} Because the reprojected coordinate $\hat{p}_t$ is generally non-integer, differentiable bilinear sampling is used to interpolate the corresponding pixel value from the source image $I_{t-1}$ and synthesize the reconstructed target image $\hat{I}_t$:
	\begin{equation}
		\hat{I}_t(\hat{p}_t) = \text{Bln}(I_{t-1}, p_{t-1})
	\end{equation}
	This view synthesis procedure establishes a differentiable connection between the estimated depth and pose parameters and the resulting photometric reprojection error, as shown in Fig.~\ref{fig:view_synthesis}.
	
	The photometric reconstruction error is defined as
	\begin{equation}
		L_{pd} = \| I_t - \hat{I}_t \|
	\end{equation}
	which measures the discrepancy between the observed target frame and its reconstructed counterpart. Within the proposed \textbf{Predict--Evaluate--Correct} loop, this error serves as the primary self-supervised feedback signal. Minimizing $L_{pd}$ drives the joint optimization of the depth parameters $\Phi$ and the pose parameters $\Omega$, thereby encouraging the model to learn scene geometry and relative ego-motion that are consistent with the incoming video stream.
	
	\subsubsection{Structural Similarity Loss and Edge-Aware Smoothness}
	The model is optimized by minimizing the photometric reprojection error between the observed target frame $I_t$ and its reconstructed counterpart $\hat{I}_t$. Following common self-supervised monocular depth estimation practice, this photometric error consists of a pixel-wise reconstruction term $L_{pd}$ and a structural similarity term $L_{ps}$:
	\begin{equation}
		L_{p} = \alpha L_{ps} + (1 - \alpha) L_{pd}
		\label{eqn_PrcErr}
	\end{equation}
	where $\alpha = 0.85$ balances structural consistency against pixel-level fidelity. The structural similarity index measure (SSIM) \cite{wang2004image} is used to capture perceptual structural consistency between the target image and the reconstructed image:
	\begin{equation}
		L_{ps} = \frac{1}{2}\left[ 1 - \mathrm{SSIM}(I_t, \hat{I}_t)\right]
	\end{equation}
	For two corresponding image patches $x$ and $y$, the SSIM term is defined as
	\begin{equation}
		\mathrm{SSIM}(x, y) = \frac{(2\mu_x\mu_y + C_1)(2\sigma_{xy} + C_2)}{(\mu_x^2 + \mu_y^2 + C_1)(\sigma_x^2 + \sigma_y^2 + C_2)}
	\end{equation}
	where $\mu_x$ and $\mu_y$ denote the local means, $\sigma_x^2$ and $\sigma_y^2$ are the local variances, $\sigma_{xy}$ is the local covariance, and $C_1, C_2$ are small constants to ensure numerical stability. This metric is particularly robust to illumination changes common in outdoor driving scenarios, ensuring that depth estimation relies on fundamental object geometry rather than absolute color intensity.
	
	Furthermore, to regularize the predicted depth map in textureless or weakly textured regions, an edge-aware smoothness term is introduced:
	\begin{equation}
		L_{s} = \sum_{p_t} \left( |\partial_x D_t(p_t)| e^{-|\partial_x I_t(p_t)|} + |\partial_y D_t(p_t)| e^{-|\partial_y I_t(p_t)|} \right)
	\end{equation}
	This term encourages the estimated depth map $D_t$ to remain smooth in locally homogeneous regions while allowing sharp depth changes at object boundaries. Here, $\partial_x$ and $\partial_y$ denote the horizontal and vertical spatial gradients, respectively. By weighting the depth gradients with image-gradient-dependent exponential factors, the formulation suppresses oversmoothing across strong visual edges.
	
	Because the depth network produces predictions at multiple spatial scales, the overall training objective is defined as a multi-scale loss:
	\begin{equation}
		L_{\sum} = \sum_s \left( \lambda_s L_p^s + \beta L_s^s \right)
	\end{equation}
	where $L_p^s$ and $L_s^s$ denote the photometric and smoothness terms computed at scale $s$, respectively. The coefficient $\lambda_s$ controls the contribution of the photometric loss at each scale, while $\beta$ is a scale-independent weight for the smoothness term. In this work, $\lambda_s$ is defined as $\lambda_s = 1/2^s$, where $s \in \{0,1,2,\ldots\}$.
	
	\subsection{Lightweight Design}
	
	\subsubsection{MobileNetV3-Small as the Shared Encoder}
	To support real-time inference and online adaptation on embedded platforms, the backbone must maintain a low parameter count and a small computational footprint. For this reason, \textbf{MobileNet}V3-Small is adopted as the shared encoder for both the depth and pose networks, as illustrated in Fig.~\ref{fig:arch_overview}.
	
	Several properties of \textbf{MobileNet}V3-Small make it suitable for the proposed \textbf{OAL} framework:
	\begin{enumerate}[label=(\arabic*), leftmargin=*]
		\item \textbf{Depthwise separable convolutions:} A standard convolution is factorized into a depthwise spatial convolution and a pointwise convolution, which substantially reduces both parameter count and computational cost.
		\item \textbf{Squeeze-and-Excitation (SE) modules:} These built-in lightweight channel-attention units adapt feature responses dynamically and improve representational efficiency without introducing heavy external attention blocks.
		\item \textbf{h-swish activation:} This activation function provides a computationally efficient approximation to swish and is well suited to efficient inference on edge hardware.
	\end{enumerate}
	
	Compared with a standard \textbf{ResNet}18 backbone \cite{he2016deep}, \textbf{MobileNet}V3-Small reduces the parameter count by approximately 75\%. Although this lightweight design may incur a modest sacrifice in static offline accuracy, it provides an efficient backbone that is better aligned with the real-time and continual adaptation requirements of the proposed framework.
	
	\subsubsection{Gated Cross-scale Additive Fusion}
	Standard \textbf{U-Net}-style architectures typically use dense concatenation to fuse encoder and decoder feature maps. Although this strategy is effective in high-capacity networks, it can be suboptimal for compact backbones such as \textbf{MobileNet}V3-Small, where shallow encoder features are more susceptible to high-frequency noise. In such cases, indiscriminate skip fusion may propagate noisy low-level activations into the decoder and interfere with depth reconstruction. This limitation is particularly pronounced in lightweight models operating under restricted representational budgets.
	
\begin{figure}[htbp]
		\centering
		\includegraphics[width=0.99\linewidth]{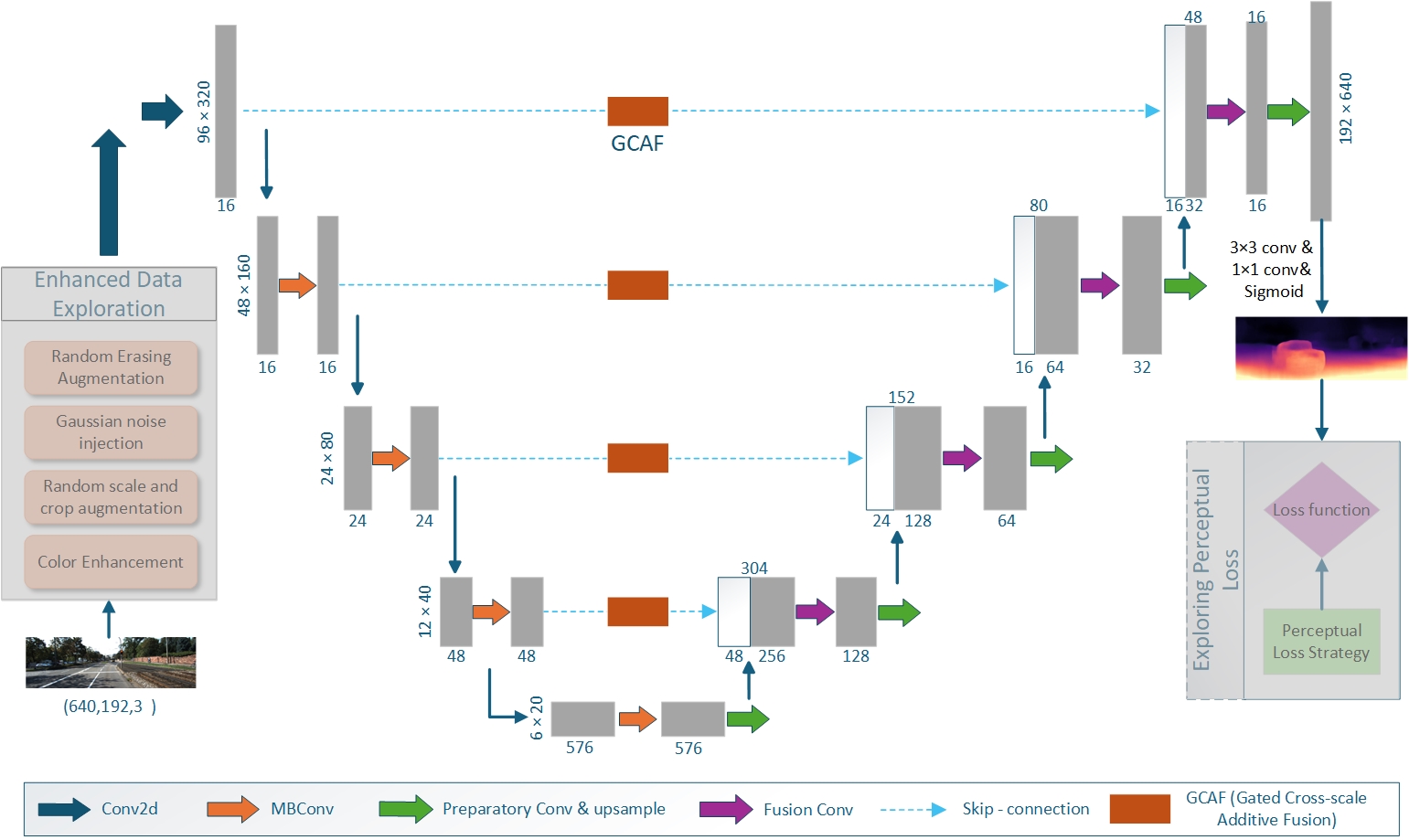}
		\caption{Detailed architecture of the proposed model incorporating the Gated Cross-scale Additive Fusion (GCAF). The GCAF modules are embedded within the skip connections, acting as dynamic filters to suppress high-frequency noise from the shallow encoder layers before fusion.}
		\label{fig:gcaf_network}
	\end{figure}

	To address this issue, a \textbf{Gated Cross-scale Additive Fusion (GCAF)} mechanism is introduced to perform top-down, semantics-guided filtering over the skip connections, as illustrated in Fig.~\ref{fig:gcaf_network}.
	
	Let $F_l^d$ denote the upsampled decoder feature map at stage $l$, and let $F_l^e$ denote the corresponding encoder feature map. Instead of concatenating these features directly, the proposed GCAF module first computes a channel-wise gating vector $g_l$ from the semantic context provided by $F_l^d$:
	\begin{equation}
		g_l = \sigma\!\left( W_2 \, \delta\!\left( W_1 \, \mathrm{AvgPool}(F_l^d) \right) \right)
	\end{equation}
	where $\mathrm{AvgPool}(\cdot)$ denotes global average pooling, $\delta(\cdot)$ is the ReLU activation function, $\sigma(\cdot)$ is the Sigmoid activation function, and $W_1$ and $W_2$ are the weight matrices of a lightweight two-layer multilayer perceptron (MLP). The resulting gate vector satisfies $g_l \in [0,1]^C$ and measures the importance of each encoder channel conditioned on the decoder semantics.
	
	The fused feature map is then defined in residual additive form as
	\begin{equation}
		F_l = F_l^d + \mathbf{g}_l \odot \mathcal{P}(F_l^e)
	\end{equation}
	where $\odot$ denotes channel-wise Hadamard multiplication, and $\mathcal{P}(\cdot)$ denotes a $1 \times 1$ projection convolution for channel alignment.
	
	This formulation allows the decoder to suppress noisy or task-irrelevant encoder responses when the corresponding gate values are small, while preserving structurally useful information. Compared with dense concatenation, the proposed fusion strategy follows a \textbf{subtractive filtering} principle rather than indiscriminate feature accumulation. As a result, it produces a cleaner intermediate representation that is beneficial for both static inference and subsequent online adaptation.
	
	The key distinction between the proposed GCAF module and conventional attention mechanisms, such as Convolutional Block Attention Module (\textbf{CBAM}) \cite{woo2018cbam}, lies in the direction of feature modulation and the receptive-field context used to generate the gating signal. Standard attention modules inserted in shallow encoder layers operate mainly in a bottom-up manner. Because they rely on relatively local feature responses, they may be less effective at separating true geometric boundaries from high-frequency background textures in lightweight backbones.
	
	By contrast, GCAF explicitly adopts top-down semantic guidance. The gating vector $\mathbf{g}_l$ is generated exclusively from the decoder feature $F_l^d$, which carries more condensed semantic information after passing through the network bottleneck. This design allows higher-level decoder semantics to regulate the incoming low-level encoder feature $F_l^e$ before fusion. As a result, GCAF acts as a top-down \textbf{subtractive filtering} mechanism that suppresses task-irrelevant high-frequency textures while preserving structurally useful information for depth reconstruction. This property is particularly beneficial in compact networks, where indiscriminate feature accumulation can destabilize optimization.
	
	For cross-scale fusion, the same gating mechanism is also applied to aggregate adjacent decoder features, which further improves contextual consistency without introducing substantial computational overhead.
	
	\subsection{Online Adaptive Learning}
\begin{figure}[!htbp]
		\centering
		\includegraphics[width=0.85\linewidth]{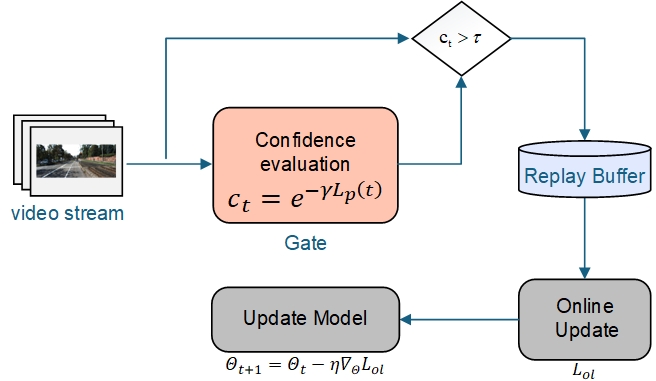}
		\caption{Workflow of the proposed OAL framework. High-confidence observations are selected for online updates, while EWC regularization constrains parameter drift to preserve prior knowledge.}
		\label{fig:ocl_flow}
	\end{figure}

	As illustrated in Fig.~\ref{fig:arch_overview} and Fig.~\ref{fig:ocl_flow}, the proposed \textbf{OAL} framework performs continuous online adaptation through a closed-loop \textbf{Predict--Evaluate--Correct} mechanism, consisting of confidence-based sample selection, replay-buffer-based mini-batch construction, and EWC-constrained online optimization.
	
	\subsubsection{Active Sample Selection via Confidence Evaluation}
	A central challenge in online self-supervised adaptation is that not every incoming frame provides a reliable learning signal. Dynamic objects, occlusions, and imaging artifacts can corrupt the photometric supervision and make direct frame-by-frame updates unstable. In the proposed framework, this issue is handled in the \textbf{Evaluate} stage of the \textbf{Predict--Evaluate--Correct} loop through a confidence-based gating mechanism.
	
	Let $L_p(t)$ denote the photometric reprojection error at time step $t$, as defined in Eqn.~(\ref{eqn_PrcErr}). A confidence score $c_t$ is then defined as
	\begin{equation}
		c_t = e^{-\gamma \cdot L_p(t)}
	\end{equation}
	where $\gamma$ is a sensitivity hyper-parameter that controls how strongly the confidence score responds to the reprojection error. This score reflects the reliability of the current self-supervised observation: a larger $c_t$ indicates that the current frame is more consistent with the predicted depth and relative pose under the rigid reconstruction model.
	
	A binary gating function is then used to determine whether the current frame should be admitted for online learning:
	\begin{equation}
		\mathrm{g}_t =
		\begin{cases}
			1, & \text{if } c_t > \tau \\
			0, & \text{otherwise}
		\end{cases}
	\end{equation}
	where $\tau$ is a predefined confidence threshold. Only frames satisfying $c_t > \tau$ are stored in the first-in-first-out (FIFO) replay buffer $\mathcal{M}$, so that unreliable inputs such as dynamic foreground objects or severe imaging artifacts are filtered out before parameter updates.
	
	\subsubsection{Incremental Online Optimization}
	Once the replay buffer $\mathcal{M}$ contains a sufficient number of reliable samples, the \textbf{Correct} stage of the online adaptation loop is activated. Model parameters are then updated using the \textbf{AdamW} optimizer. Unlike offline training, which seeks global convergence over a fixed dataset, the online optimization process aims to progressively reduce the local domain gap while maintaining update stability.
	
	The parameter update at time step $t$ is defined as
	\begin{equation}
		\Theta_{t+1} \leftarrow \Theta_t - \eta \cdot \nabla_{\Theta} L_{ol}(\mathcal{B})
	\end{equation}
	where $\eta$ is a reduced learning rate (typically $10^{-6}$) used to prevent catastrophic parameter drift, $\mathcal{B} \subset \mathcal{M}$ denotes the sampled mini-batch, and $L_{ol}$ denotes the composite online loss over both depth parameters $\Phi$ and pose parameters $\Omega$, with an additional \textbf{EWC}-based regularization term to preserve previously learned structural priors.
	
	\subsubsection{Stability Preservation via Elastic Weight Consolidation}
	Online adaptation inherently involves a stability--plasticity trade-off. While minimizing the current online loss helps the model adapt to the incoming video stream, unconstrained updates may drive the parameters away from the structural priors acquired during offline training. To address this issue, \textbf{Elastic Weight Consolidation} (\textbf{EWC}) is introduced as a regularization mechanism for stable online adaptation.
	
	Before deployment, the importance of each model parameter is estimated using the Fisher Information Matrix (FIM). In particular, parameters associated with the \textbf{high-Fisher information space} are interpreted as carrying more important global structural knowledge and therefore should be more strongly preserved during online updates. In practice, the diagonal of the FIM is approximated from the gradients of the offline training objective $L_{of}$ evaluated over a subset of the source dataset $\mathcal{D}_{s}$:
	\begin{equation}
		\breve{F} = \mathbb{E}_{x \sim \mathcal{D}_{s}} \left[ \left( \frac{\partial}{\partial \Theta} \log L_{of}(x,\Theta)\right)^2 \right]_{\Theta=\Theta^*}
	\end{equation}
	where $\breve{F}$ denotes the vectorized diagonal approximation of the FIM, such that $\breve{F}_i$ corresponds to the importance of parameter $\theta_i \in \Theta$, and $\Theta^*$ denotes the optimal parameters obtained during offline pre-training. A large $\breve{F}_i$ indicates that $\theta_i$ is more critical to the learned global structural representation and therefore belongs to the \textbf{high-Fisher information space}, whereas a small $\breve{F}_i$ places the parameter in a more flexible \textbf{low-Fisher} subspace.
	
	During online adaptation, a quadratic regularization term is introduced to constrain the parameter trajectory around $\Theta^*$. The EWC loss is defined as
	\begin{equation}
		L_E(\Theta) = \frac{1}{2} \sum_i \breve{F}_i (\theta_i - \theta_i^*)^2
		= \frac{1}{2} [\breve{F} \odot (\Theta - \Theta^*)]^T (\Theta - \Theta^*)
	\end{equation}
	where $\odot$ denotes element-wise multiplication. This term anchors each parameter to its offline value $\theta_i^*$ with a stiffness proportional to its estimated importance $\breve{F}_i$. As a result, the online optimizer is encouraged to preserve parameters in the \textbf{high-Fisher information space} while adapting primarily through the more flexible \textbf{low-Fisher} subspace.
	
	\subsubsection{Total Online Optimization Objective}
	
	The composite online objective $L_{ol}$ is designed to balance three complementary goals: 1) adaptation to the current target stream, 2) geometric regularization of the predicted depth map, and 3) preservation of previously learned structural knowledge. It is defined as
	\begin{equation}
		L_{ol} = \underbrace{\lambda_{pd} L_p(\mathcal{B})}_{\text{Plasticity}} + \underbrace{\lambda_{sm} L_s(\mathcal{B})}_{\text{Smoothness}} + \underbrace{\lambda_{ec} L_E(\Theta)}_{\text{Stability}}
	\end{equation}
	where $\mathcal{B}$ denotes the current mini-batch sampled from the replay buffer, $L_p$ drives target-domain adaptation, $L_s$ regularizes the predicted depth geometry, and $L_E$ constrains parameter drift through EWC, with $\lambda_{pd}$, $\lambda_{sm}$, and $\lambda_{ec}$ controlling their relative contributions. From an optimization perspective, the EWC term adds a convex quadratic component centered at $\Theta^*$, effectively restricting updates within a dynamic \textbf{Trust Region}. In particular, the update magnitude for each parameter is modulated by its Fisher importance:
	\begin{equation}
		\|\Delta \theta\| \propto \frac{1}{\breve{F}_\theta + \epsilon} \|\nabla_{\theta} L_p\|
	\end{equation}
	where $\epsilon$ is a small positive constant introduced for numerical stability. This bounded-update behavior helps keep the optimization trajectory near the manifold of valid structural representations, thereby reducing the risk of degenerate solutions under noisy self-supervised signals.
	
	\subsection{Convergence Analysis of Online Optimization}
	The stability of online adaptation depends on the local geometry of the optimization objective: naive fine-tuning on the target stream induces a highly non-convex landscape, whereas the proposed \textbf{OAL} framework introduces an \textbf{EWC} regularization term that penalizes deviations from the offline optimum $\Theta^*$ and thereby constrains the update dynamics. The update rule at time step $t$ is written as
	\begin{equation}
		\begin{aligned}
			\Theta_{t+1} = \Theta_t - \eta \nabla_{\Theta} \Big[
			&\lambda_{pd} L_{p}(\Theta_t) \\
			&+ \frac{\lambda_{ec}}{2} [\breve{F} \odot (\Theta_t - \Theta^*)]^T (\Theta_t - \Theta^*)
			\Big]
		\end{aligned}
	\end{equation}
	where $\breve{F}$ denotes the diagonal Fisher information used to approximate the local curvature of the offline training loss. Geometrically, this regularization term superimposes a \textbf{convex quadratic potential} centered at the offline optimum $\Theta^*$. Because $\breve{F}$ is positive semi-definite, the regularized objective exhibits increased curvature, especially along directions that are important to previously learned structural knowledge.
	
	Theoretically, this constraint implies that the parameter update can be bounded within a dynamic \textbf{Trust Region}:
	\begin{equation}
		\| \Theta_{t+1} - \Theta^* \| \le \left(1 - \eta \lambda_{ec} \rho_{\min}(\breve{F})\right) \| \Theta_t - \Theta^* \| + \eta \lambda_{pd} \|\nabla L_{p}\|
	\end{equation}
	where $\rho_{\min}(\breve{F})$ denotes the minimum diagonal value of the Fisher information term. Consequently, the parameter trajectory remains confined near the offline baseline, allowing the model to adapt to new environmental cues without catastrophic deviation from its core structural priors.
	
	% ==========================================
	% SECTION IV: EXPERIMENTS AND RESULTS
	% ==========================================
	\section{Experiments and Results}
	\label{sec:experiments}
	
	\subsection{Experimental Setup}
	
	\subsubsection{Dataset and Data Preprocessing}
	To evaluate the performance and adaptability of the proposed lightweight monocular depth estimation framework \textbf{OAL}, the experiments were divided into an offline training phase for baseline establishment and an online continual adaptation phase for robustness evaluation under changing conditions.
	
	The primary evaluation was conducted on the standard \textbf{KITTI} dataset \cite{geiger2012are}, using the standard \textbf{Eigen split} with 39,810 training frames from 38 subsequences and 697 validation frames. Raw KITTI sequences excluded from offline training were used for the initial online adaptation phase, while \textbf{NuScenes} \cite{caesar2020nuscenes} was introduced in the zero-shot online adaptation setting to evaluate cross-domain robustness under different camera intrinsics and geographic environments.
	
	During preprocessing, all input images were resized to $640 \times 192$ to balance spatial detail and computational efficiency. Camera calibration followed the official intrinsic matrix $K$ to ensure geometric consistency during view synthesis. In \textbf{KITTI}, the ground-truth depth maps were obtained from projected LiDAR point clouds and were used \textbf{only} for evaluation during testing; they were not used during training. All preprocessing settings were kept fixed across experiments, while the optimization-related settings are described in the following subsection.
	
	\subsubsection{Training Settings and Data Augmentation}
	All experiments were implemented in PyTorch on a single NVIDIA RTX 4060 GPU. Unless otherwise specified, offline training used 50 epochs, a batch size of 12, and Adam with an initial learning rate of $10^{-4}$; Monodepth2-style data augmentation, including random horizontal flipping and color jittering, was applied during offline training but disabled during online adaptation, where \textbf{AdamW} with a reduced learning rate of $\eta = 10^{-6}$ was used to preserve temporal consistency and limit catastrophic parameter drift.
	
	\subsubsection{Evaluation Metrics and Experimental Phase Design}
	
	\begin{figure}[htbp]
		\centering
		\includegraphics[width=0.95\linewidth]{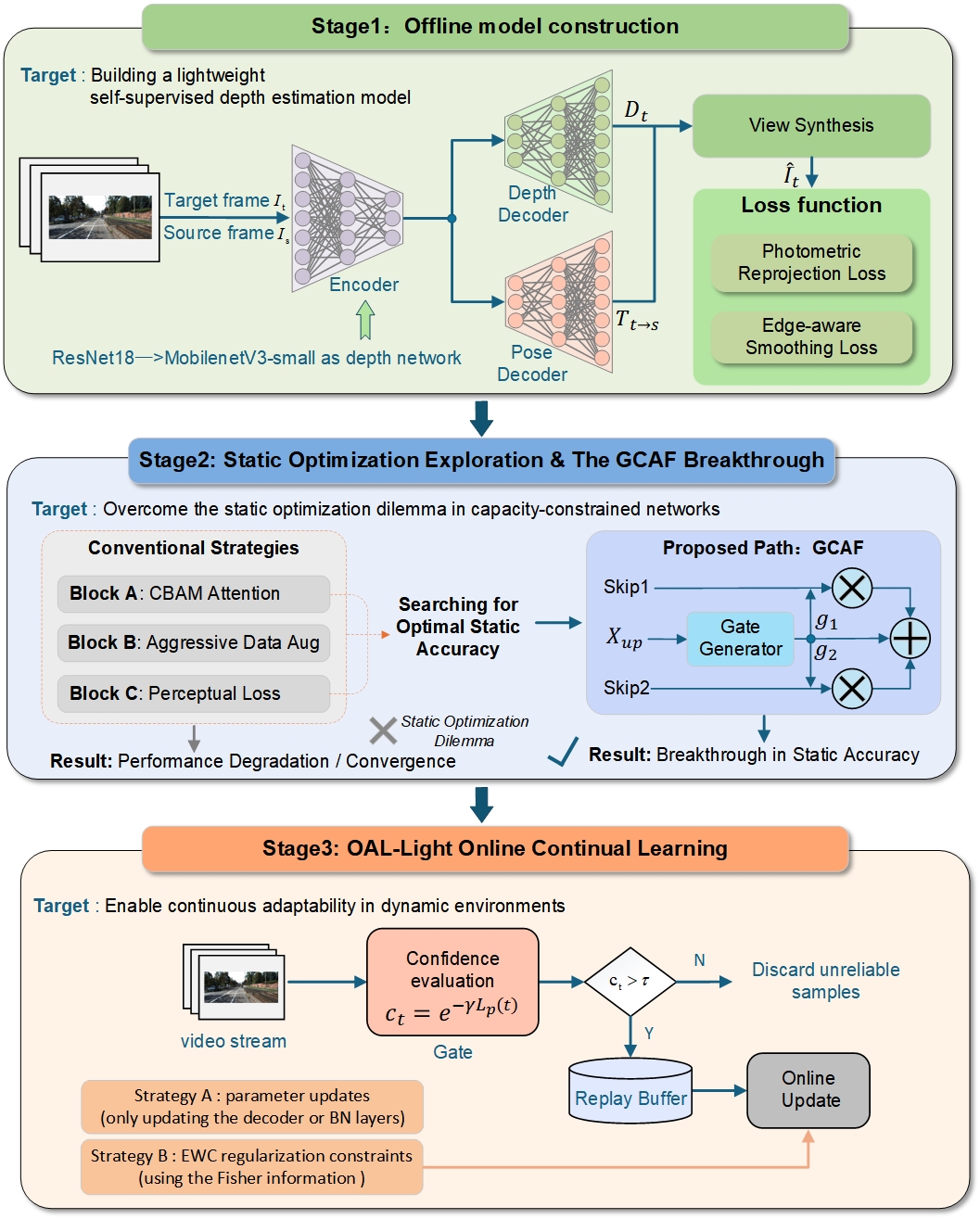}
		\caption{\textbf{Schematic overview of the proposed OAL research framework.} The methodology proceeds in three stages. \textbf{Stage 1} establishes the lightweight baseline model under self-supervised monocular depth learning. \textbf{Stage 2} evaluates the static optimization dilemma of lightweight architectures and examines how the proposed GCAF mechanism improves representation efficiency under restricted model capacity. \textbf{Stage 3} evaluates continuous online adaptation through confidence-driven sample filtering, replay-based updates, and EWC-constrained optimization to verify stable domain alignment during deployment.}
		\label{fig:research_framework}
	\end{figure}
	
	To evaluate predictive performance under the proposed framework (Fig.~\ref{fig:research_framework}), standard monocular depth estimation metrics are adopted, including Abs Rel, Sq Rel, RMSE, RMSE log, and threshold accuracies ($\delta < 1.25, 1.25^2, 1.25^3$), following the standard protocol in \cite{eigen2014depth}.
	
	The overall experimental protocol follows the three-stage framework illustrated in Fig.~\ref{fig:research_framework}. It first establishes the lightweight baseline, then evaluates the static optimization dilemma and the contribution of \textbf{GCAF}, and finally assesses whether the proposed \textbf{OAL} framework improves robustness under in-domain and cross-domain shifts.
	
	\subsection{Improving Representation Efficiency with GCAF}
	
	To establish a computation-efficient foundation for subsequent online adaptation, the static representational efficiency of the lightweight backbone was first examined. Because lightweight encoders such as \textbf{MobileNet}V3-Small are more susceptible to high-frequency noise in shallow layers, the proposed \textbf{GCAF} mechanism introduces top-down, semantics-guided skip filtering to suppress task-irrelevant responses while preserving structurally useful information for depth reconstruction.
	
	As shown in Table~\ref{tab:sota}, this structural refinement yields a consistent improvement and establishes a stronger lightweight baseline for later online adaptation. The proposed \textbf{Gated} variant built on \textbf{MobileNet}V3-Small achieves an Absolute Relative (Abs Rel) error of 0.126 and a Squared Relative (Sq Rel) error of 1.021 on the \textbf{KITTI} Eigen split, improving over the naive \textbf{M1} baseline (Abs Rel 0.131, Sq Rel 1.054). Standard metrics including RMSE, RMSE log, and threshold accuracies ($\delta_1$, $\delta_2$, and $\delta_3$) are also reported in Table~\ref{tab:sota}, showing that the proposed gating strategy improves not only relative error but also the overall reliability of the predicted depth map. These results indicate that, for capacity-constrained models, selective subtractive filtering is more effective than indiscriminate additive feature accumulation.
	
	\begin{table*}[htbp]
		\centering
		\renewcommand{\arraystretch}{1.2} 
		\caption{Comparison with state-of-the-art lightweight and standard monocular depth estimation methods on the KITTI Eigen split. All methods are evaluated with an input resolution of $640 \times 192$.}
		\label{tab:sota}
		
		\resizebox{\textwidth}{!}{%
			\begin{tabular}{l c c c c c c c c}
				\toprule
				\textbf{Method} & \textbf{Params(M)} & \textbf{AbsRel}$\downarrow$ & \textbf{SqRel}$\downarrow$ & \textbf{RMSE}$\downarrow$ & \textbf{RMSElog}$\downarrow$ & $\boldsymbol{\delta_1}\uparrow$ & $\boldsymbol{\delta_2}\uparrow$ & $\boldsymbol{\delta_3}\uparrow$ \\
				\midrule
				Monodepth2~\cite{godard2019digging} & 14.33 & 0.115 & 0.903 & 4.863 & 0.193 & 0.877 & 0.959 & 0.981 \\
				\midrule
				Lite-HR-Depth~\cite{zhang2023lite} & 3.10 & 0.116 & 0.845 & 4.841 & 0.190 & 0.866 & 0.957 & 0.982 \\
				R-MSFM6~\cite{zhang2023lite} & 3.80 & 0.112 & 0.806 & 4.704 & 0.191 & 0.878 & 0.960 & 0.981 \\
				Lite-Mono~\cite{zhang2023lite} & 3.10 & \textbf{0.107} & \textbf{0.765} & \textbf{4.561} & \textbf{0.183} & \textbf{0.886} & \textbf{0.963} & \textbf{0.983} \\
				\midrule
				\textbf{M1} (Baseline) & 3.59 & 0.131 & 1.054 & 5.233 & 0.209 & 0.847 & 0.951 & 0.978 \\
				\textbf{Gated} (M1+GCAF) & 3.76 & 0.126 & 1.021 & 5.120 & 0.204 & 0.856 & 0.954 & 0.979 \\
				\bottomrule
			\end{tabular}%
		}
		
		\begin{flushleft}
			\textit{Note:} $\uparrow$/$\downarrow$ indicate whether higher/lower values are better. $\delta_1$, $\delta_2$, $\delta_3$ correspond to thresholds $1.25$, $1.25^2$, and $1.25^3$, respectively. Best results are highlighted in bold.
		\end{flushleft}
	\end{table*}
	
	Although recent hybrid lightweight architectures such as Lite-Mono achieve lower static depth error, they generally incur higher memory access cost and inference complexity. By contrast, the purely convolutional \textbf{MobileNet}V3-Small backbone provides a lightweight baseline better suited to continual online correction.
	
	\subsection{The Static Optimization Dilemma: Evidence from Negative Results}
	
	Despite the gain brought by \textbf{GCAF}, further attempts to improve the lightweight model through conventional offline enhancement strategies reveal a clear performance bottleneck: as summarized in Table~\ref{tab:ablation}, additional attention, perceptual loss, and stronger augmentation fail to provide a consistent gain and may even degrade performance, indicating that static offline enhancement is no longer reliably effective in this compact regime.
	
	\begin{table}[htb]
		\centering
		\caption{\textit{Static Optimization Dilemma} in lightweight models. Conventional offline enhancement strategies often fail to improve performance, whereas GCAF provides a consistent gain.}
		\label{tab:ablation}
		\renewcommand{\arraystretch}{1.1}
		\setlength{\tabcolsep}{3.5pt}
		\begin{tabular}{p{0.32\columnwidth}cccc}
			\toprule
			\textbf{Method} & \textbf{Abs Rel} & \textbf{Sq Rel} & \textbf{RMSE} & $\boldsymbol{\delta_1}$ \\
			\midrule
			\textbf{M1} (Baseline) & 0.131 & 1.054 & 5.233 & 0.847 \\
			\midrule
			M1 + CBAM & 0.133 & 1.060 & 5.270 & 0.845 \\
			M1 + Perceptual Loss ($\lambda=0.005$) & 0.134 & 1.002 & 5.225 & 0.836 \\
			M1 + Gaussian Noise & 0.132 & 1.041 & 5.251 & 0.847 \\
			\midrule
			M1 + Random Erasing & 0.132 & 1.092 & 5.274 & 0.847 \\
			M1 + Random Scaling \& Cropping & 0.199 & 1.624 & 7.187 & 0.676 \\
			\midrule
			\textbf{Gated} (M1+GCAF) & \textbf{0.126} & \textbf{1.021} & \textbf{5.120} & \textbf{0.856} \\
			\bottomrule
		\end{tabular}
		
		\vspace{2pt}
		\raggedright
		\footnotesize
		$\delta_1$ denotes the accuracy threshold $\delta < 1.25$.
	\end{table}
	
\begin{figure*}[!htbp]
		\centering
		\includegraphics[width=0.99\linewidth]{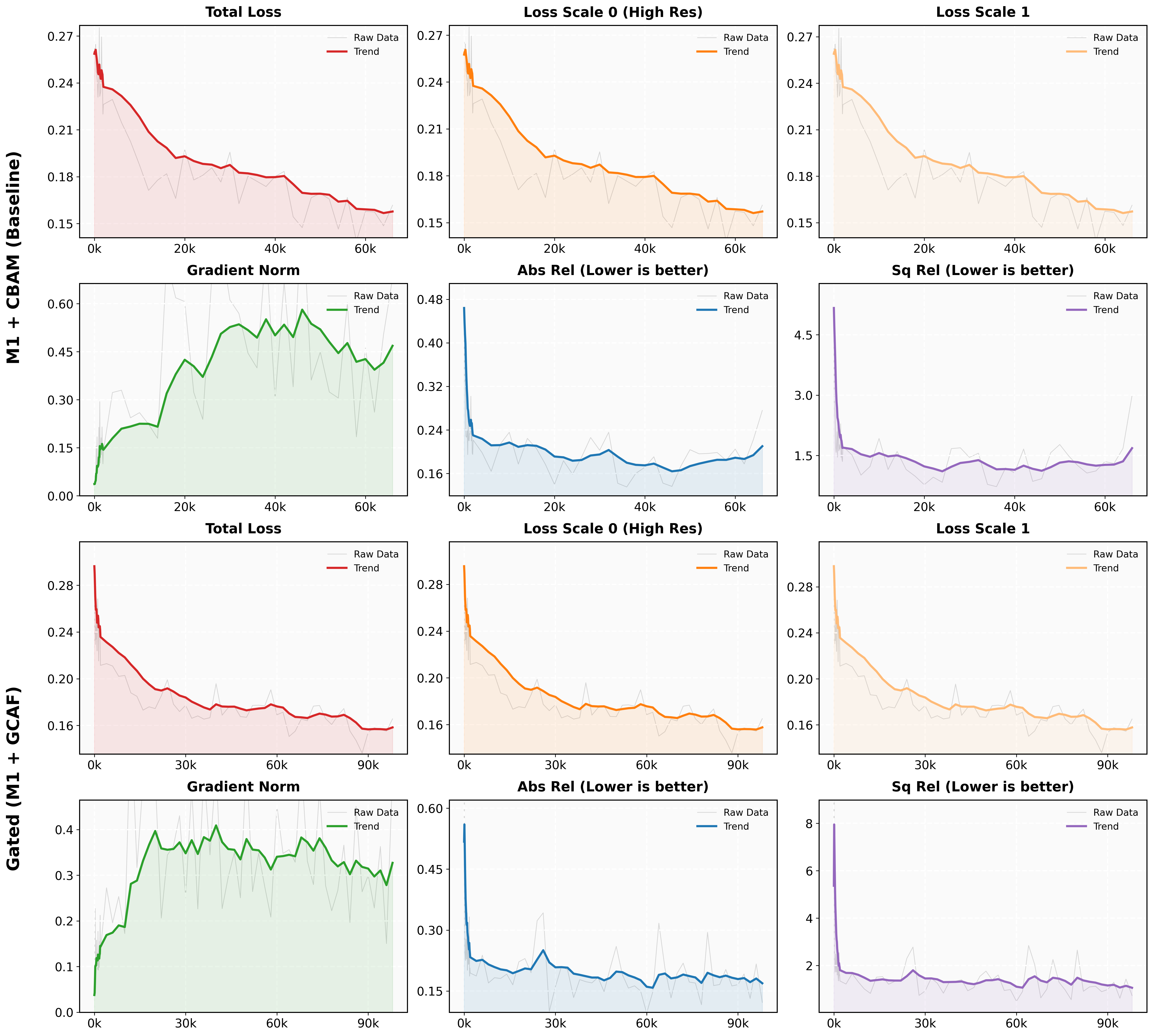}
		\caption{\textbf{Training convergence and gradient telemetry illustrating the static optimization dilemma.} \textbf{(Top)} The baseline model enhanced with \textbf{CBAM} exhibits unstable optimization behavior, characterized by larger oscillation in gradient norm during training. \textbf{(Bottom)} The proposed \textbf{GCAF} structure suppresses task-irrelevant noise in skip fusion, leading to smoother convergence and improved training stability.}
		\label{fig:gradient_dilemma}
	\end{figure*}

	The integration of \textbf{CBAM} does not improve the baseline model and instead slightly worsens Abs Rel from 0.131 to 0.133, likely because \textbf{MobileNet}V3-Small already contains built-in \textbf{SE} modules and therefore gains little from additional attention. This difference is also reflected in Fig.~\ref{fig:gradient_dilemma}, where the \textbf{CBAM}-enhanced baseline exhibits stronger gradient oscillation, whereas the proposed \textbf{GCAF} structure yields a more stable optimization trajectory.
	
\begin{figure*}[!htbp]
		\centering
		\includegraphics[width=0.99\linewidth]{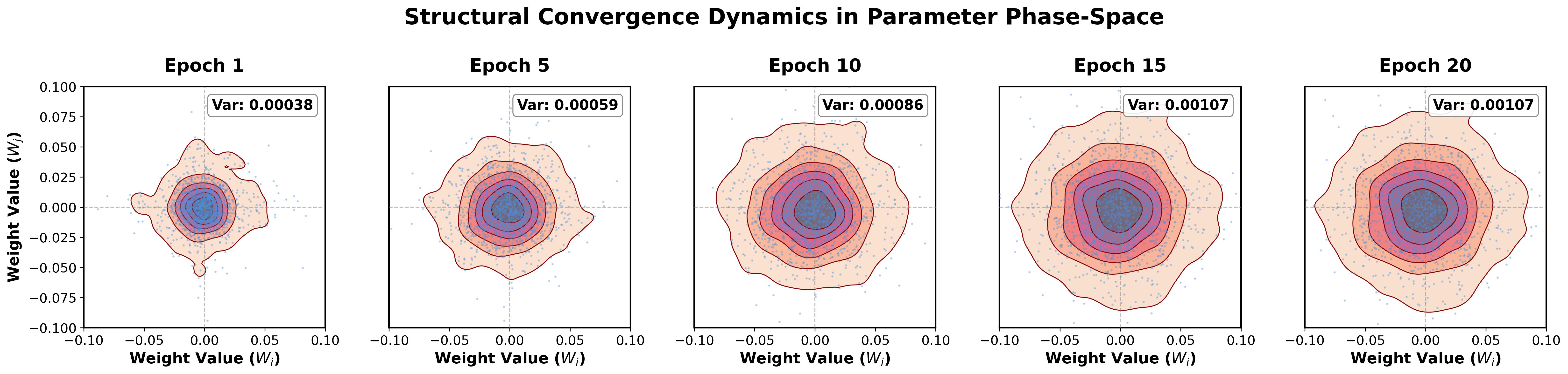}
		\caption{\textbf{Structural convergence dynamics in parameter phase-space.} The 2D kernel density contours illustrate the transition of the \textbf{GCAF} model from random initialization toward a stable functional manifold. Early epochs show parameters concentrated near zero, whereas later epochs exhibit outward expansion and eventual stabilization, indicating progressive acquisition of structural representation.}
		\label{fig:phase_space_offline}
	\end{figure*}

	The structural convergence of the \textbf{GCAF} model can be distinctly observed in the parameter phase-space (Fig.~\ref{fig:phase_space_offline}). The outward expansion and eventual stabilization of the 2D weight density empirically validate the network's successful transition from random initialization to a well-structured functional manifold.
	
\begin{figure}[!htbp]
		\centering
		\includegraphics[width=0.85\linewidth]{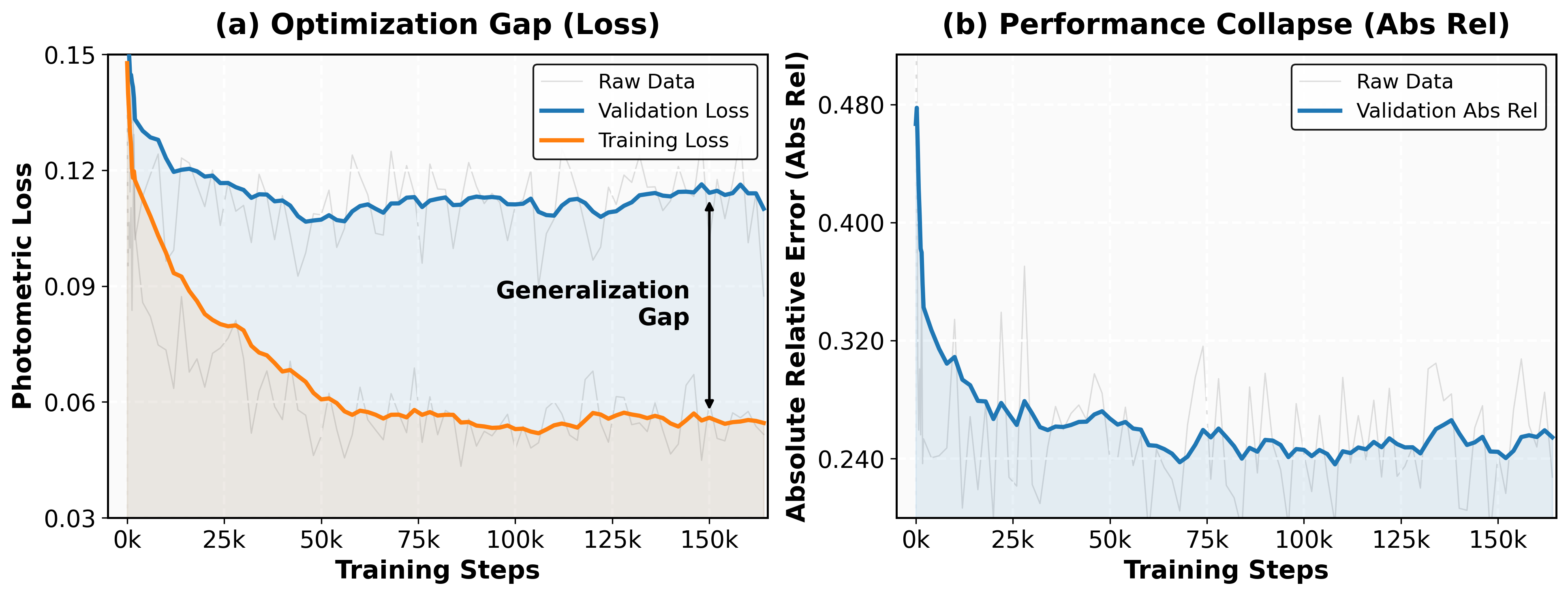}
		\caption{\textbf{Impact of aggressive geometric augmentation on a capacity-constrained model.} \textbf{(a)} A growing gap between training and validation photometric loss appears under aggressive augmentation. \textbf{(b)} The same setting leads to persistently worse validation Abs Rel, indicating degraded generalization.}
		\label{train_val_loss_curve}
	\end{figure}

	A more severe failure appears under stronger geometric augmentation: random scaling and cropping increase Abs Rel to 0.199, suggesting that compact models lack sufficient redundancy to absorb large perturbations during static offline training. This interpretation is consistent with Fig.~\ref{train_val_loss_curve}, where aggressive augmentation produces a clear train--validation gap and degraded validation Abs Rel.
	
	Taken together, these results reveal a clear \textbf{Static Optimization Dilemma}: once the representational budget of a lightweight backbone becomes saturated, further static offline enhancement is no longer a reliable path to improved deployment performance, thereby motivating the proposed \textbf{OAL} strategy for continual online adaptation.
	
	\subsection{Further Performance Improvement via Online Active Learning}
	
	Although \textbf{GCAF} improves static representational efficiency, fixed model parameters still limit the ability of the lightweight backbone to handle domain shifts after deployment; therefore, the proposed \textbf{OAL} framework is introduced to enable continual self-supervised online adaptation and to examine how online correction interacts with static structural optimization.
	
	Applied first to the \textbf{M1} + \textbf{CBAM} model, \textbf{OAL} reduces Abs Rel from 0.133 to 0.130 (Table~\ref{tab:oal_synergy}), indicating that closed-loop online feedback can partially compensate for deficiencies in offline optimization. When combined with the statically improved \textbf{Gated} (\textbf{M1}+\textbf{GCAF}) model, \textbf{OAL} yields a smaller relative gain but the best absolute performance (Table~\ref{tab:oal_synergy}), whereas the noisier \textbf{CBAM} baseline leaves a larger margin for online correction. This suggests that the statically refined \textbf{Gated} backbone removes much of the low-level structural noise and allows online adaptation to focus more directly on genuine domain alignment.

	\begin{table}[htbp]
		\centering
		\caption{Impact of \textbf{OAL} on different lightweight models. While the sub-optimal CBAM architecture exhibits a larger relative error reduction due to its higher initial noise margin, combining \textbf{OAL} with the proposed \textbf{GCAF} structure achieves the highest absolute performance (Abs Rel = 0.125).}
		\label{tab:oal_synergy}
		\renewcommand{\arraystretch}{1.12}
		\setlength{\tabcolsep}{4pt}
		\begin{tabular}{p{0.32\columnwidth}ccc}
			\toprule
			\textbf{Method} & \textbf{Abs Rel (Offline)} & \textbf{Abs Rel (OAL)} & \textbf{Gain} \\
			\midrule
			M1 + CBAM & 0.133 & 0.130 & +2.25\% \\
			\textbf{Gated} (M1+GCAF) & 0.126 & \textbf{0.125} & +0.79\% \\
			\bottomrule
		\end{tabular}
	\end{table}
	
	These results show that a better static architecture does not eliminate the need for online adaptation; instead, it allows \textbf{OAL} to focus more directly on deployment-time domain alignment rather than architectural correction.
	
	\subsection{Resolving the Stability-Plasticity Trade-off}
	The effect of the proposed \textbf{OAL} framework on the stability--plasticity trade-off is summarized in Table~\ref{tab:ewc}. When the statically optimized \textbf{Gated} model is deployed in a new target stream, the target-domain Abs Rel increases from 0.126 to 0.139 because of domain shift, whereas \textbf{OAL} reduces it to 0.132 while preserving the original-domain performance at 0.125, consistent with the intended role of \textbf{EWC} in protecting the \textbf{high-Fisher information space} while allowing adaptation through the more flexible \textbf{low-Fisher} subspace.
	
	\begin{table}[htb]
		\centering
		\caption{Comparison of stability and adaptability under different online strategies. The proposed EWC regularization effectively prevents performance degradation on the original domain while allowing adaptation to the target domain.}
		\label{tab:ewc}
		\setlength{\tabcolsep}{8pt}
		\begin{tabular}{lcc}
			\toprule
			\textbf{Online Strategy} & \textbf{Abs Rel(Original)} & \textbf{Abs Rel(Target)} \\
			\midrule
			\textbf{Gated} (Static)  & 0.126 & 0.139\\
			\textbf{Gated} + \textbf{OAL} & 0.125 & 0.132 \\
			\bottomrule
		\end{tabular}
	\end{table}
	
\begin{figure}[!htbp]
		\centering
		\includegraphics[width=1\linewidth]{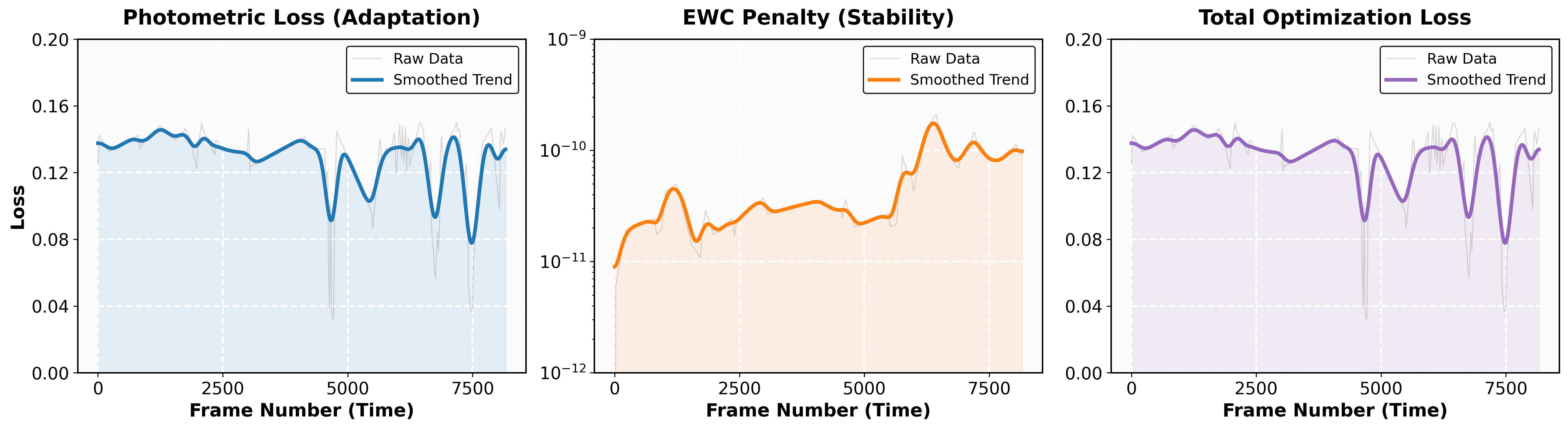}
		\caption{\textbf{Optimization dynamics during the continuous online adaptation phase.} The real-time telemetry over 8000 continuous frames illustrates the resolution of the stability-plasticity dilemma: the photometric loss (left) adapts to local domain shifts (plasticity), while the EWC penalty (middle) actively constrains catastrophic parameter drift (stability), ensuring the total optimization loss (right) converges smoothly without catastrophic forgetting.}
		\label{fig:online_metrics}
	\end{figure}

	The optimization dynamics during an 8,000-frame continuous adaptation sequence (Fig.~\ref{fig:online_metrics}) further confirm this coordinated regulation. The concurrent convergence of the photometric loss and the \textbf{EWC} penalty empirically demonstrates that the framework effectively aligns with local domain shifts while strictly bounding catastrophic parameter drift.
	
\begin{figure*}[!htbp]
		\centering
		\includegraphics[width=0.93\textwidth]{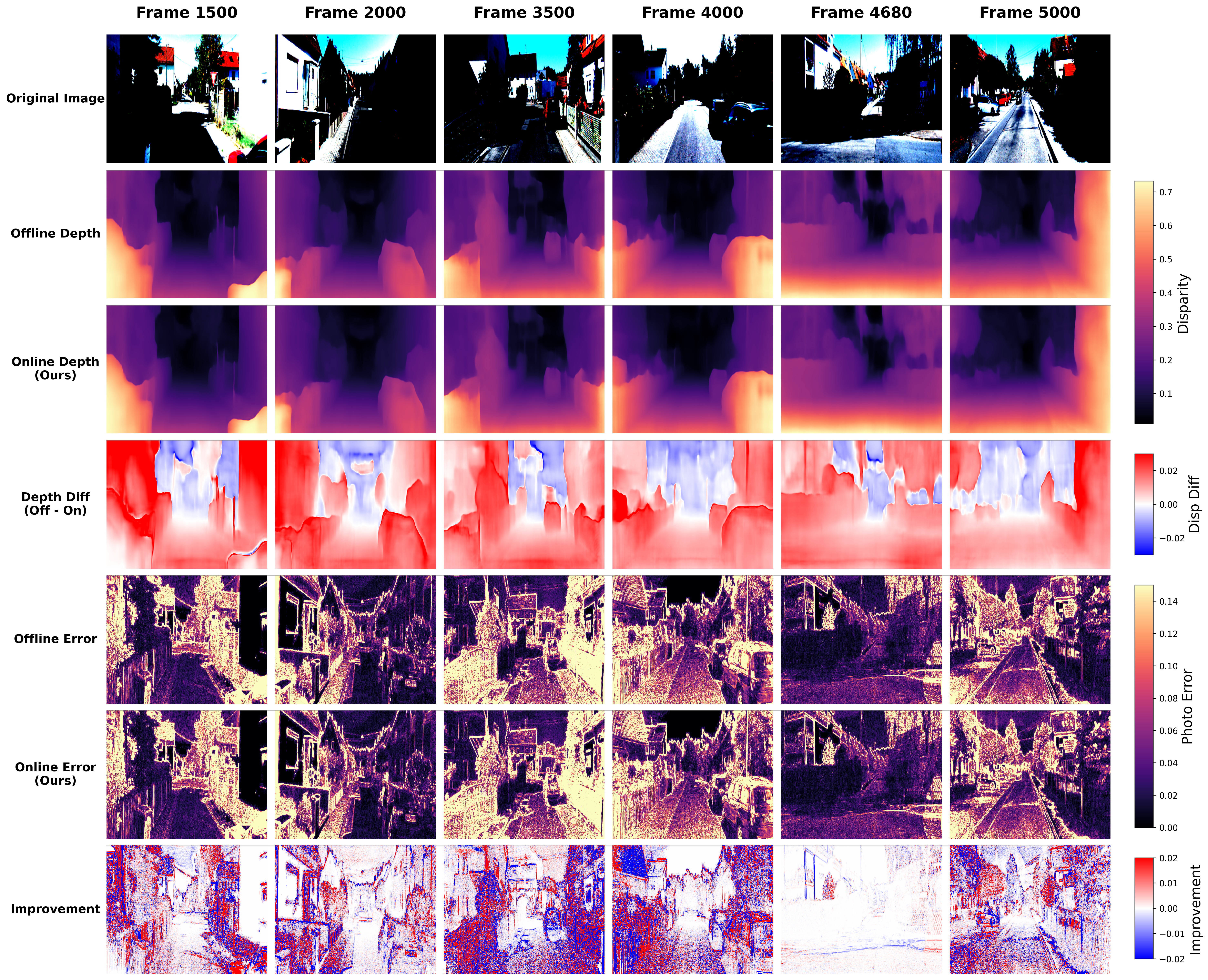}
		\caption{\textbf{Qualitative comparison between the static offline model and the proposed OAL framework.} The visual results across continuous streaming frames demonstrate that the online adapted depth maps (Online Depth) effectively suppress the high-photometric errors present in the frozen offline model (Offline Error). The Improvement mask (bottom row, blue indicates error reduction) confirms the dynamic domain alignment capability of our method.}
		\label{fig:qualitative_heatmap}
	\end{figure*}

	Fig.~\ref{fig:qualitative_heatmap} further confirms that continuous online adaptation suppresses localized photometric failures while preserving overall structural consistency.
	
	\subsection{Computational Complexity and Inference Efficiency}
	\label{sec:computational_complexity}
	
	To evaluate structural efficiency, the static M1 baseline is compared with the proposed Gated structure in terms of parameter count, MACs, FLOPs, peak VRAM consumption, and inference throughput (FPS), using an input resolution of $640 \times 192$ on a single NVIDIA RTX 4060 GPU. The results are summarized in Table~\ref{tab:computational_overhead}.
	
	\begin{table}[htbp]
		\centering
		\caption{Computational overhead and inference efficiency comparison.}
		\label{tab:computational_overhead}
		\resizebox{\linewidth}{!}{
			\begin{tabular}{lccc}
				\toprule
				Metric & M1 (Static Baseline) & Gated (Ours) & Increment ($\Delta$) \\
				\midrule
				Parameters (M)   & 3.587  & 3.763  & + 0.176 (+4.9\%) \\
				MACs (G)         & 2.641  & 2.865  & + 0.224 (+8.4\%) \\
				FLOPs (G)        & 5.282  & 5.729  & + 0.447 (+8.4\%) \\
				Peak VRAM (MB)   & 130.18 & 172.87 & + 42.69 \\
				Throughput (FPS) & 440.50 & 388.47 & - 52.03 \\
				\bottomrule
				\multicolumn{4}{p{\linewidth}}{\small \textit{Note: Throughput is measured under strict synchronized conditions after sufficient GPU warm-up.}}
			\end{tabular}
		}
	\end{table}
	
	As shown in Table~\ref{tab:computational_overhead}, the proposed GCAF mechanism imposes only a modest overhead in parameters and FLOPs while maintaining low VRAM consumption and high inference throughput. These results indicate that the purely convolutional design remains suitable for real-time perception and leaves sufficient computational headroom for online adaptation.
	
	\subsection{Cross-Domain Scale Consistency for Continuous Adaptation}
	\label{sec:scale_consistency}
	
	Maintaining scale consistency is critical in continuous monocular adaptation because repeated online updates may otherwise induce severe scale drift. To evaluate robustness against such degradation, a scale sensitivity analysis is conducted in both in-domain (KITTI) and cross-domain (NuScenes) settings.
	
	\begin{figure*}[htbp]
		\centering
		\begin{minipage}{0.48\textwidth}
			\centering
			\includegraphics[width=\linewidth]{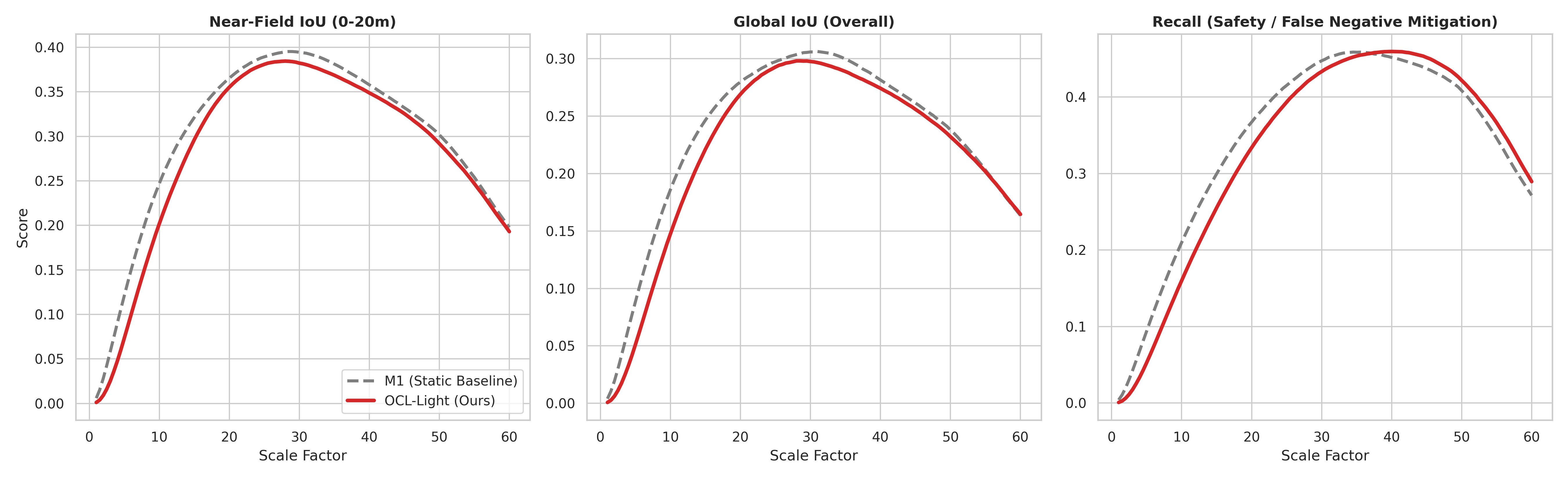}
			\caption{Scale sensitivity analysis on the in-domain KITTI dataset. The proposed framework demonstrates a stable error basin across varying scale factors, indicating robust resistance to geometric distortion during adaptation.}
			\label{fig:scale_kitti}
		\end{minipage}\hfill
		\begin{minipage}{0.48\textwidth}
			\centering
			\includegraphics[width=\linewidth]{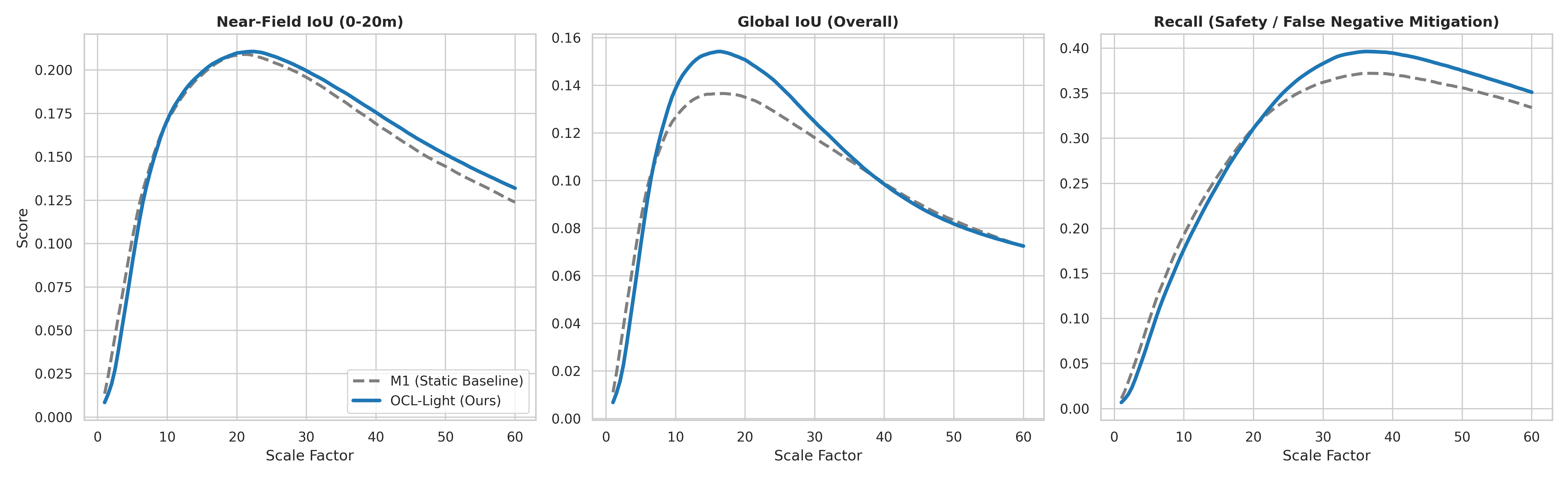}
			\caption{Cross-domain scale sensitivity analysis on the unseen NuScenes dataset. The sustained stability under severe domain shift verifies the framework's capability to prevent catastrophic scale drift during continuous deployment.}
			\label{fig:scale_nuscenes}
		\end{minipage}
	\end{figure*}
	
	As illustrated in Fig.~\ref{fig:scale_kitti} and Fig.~\ref{fig:scale_nuscenes}, the proposed framework exhibits a smoother and wider error basin than the static baseline and remains stable even under severe cross-domain shift. These results indicate that the proposed \textbf{OAL} framework preserves structural and scale consistency during continuous deployment, thereby providing a more reliable geometric basis for downstream spatial perception tasks.

	% ==========================================
	% SECTION V: DISCUSSION AND CONCLUSION
	% ==========================================
	\section{Discussion and Conclusion}
	\label{sec:discussion}
	
	\subsection{Structural Parsimony in Capacity-Constrained Architectures}
	The results indicate that strategies often effective in large-scale models, such as additional attention modules, aggressive data augmentation, or heavy semantic supervision, may instead degrade lightweight architectures. Unlike over-parameterized backbones such as \textbf{ResNet}101 \cite{he2016deep}, compact encoders such as \textbf{MobileNet}V3-Small operate under strict channel-budget constraints and are therefore more susceptible to high-frequency noise and optimization conflict.
	
	\begin{figure}[htbp]
		\centering
		\includegraphics[width=0.99\linewidth]{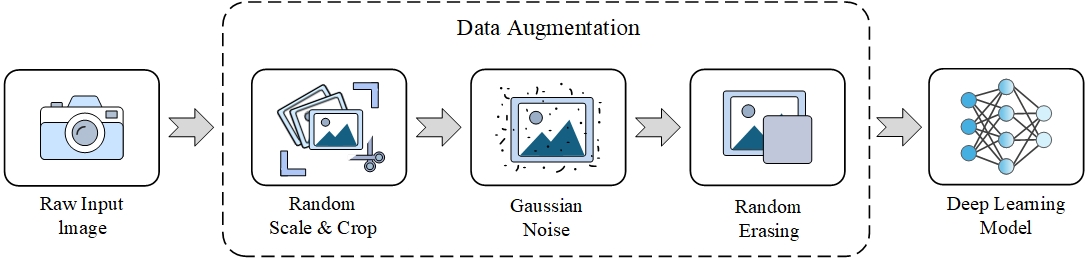}
		\caption{The aggressive data augmentation pipeline exemplifies capacity overload in lightweight networks. Forcing the compact \textbf{MobileNet}V3-Small encoder to learn invariances against complex combinations of scaling, cropping, noise, and erasing triggers optimization conflicts rather than improved generalization.}
		\label{fig:data_aug}
	\end{figure}
	
	As shown by the negative results in Table~\ref{tab:ablation}, perceptual supervision and additional complex attention do not improve the compact backbone, whereas the proposed \textbf{GCAF} mechanism provides a consistent gain. These observations suggest that, for edge-oriented networks, \textbf{structural parsimony}, rather than increased architectural complexity, offers a more suitable inductive bias for improving static efficiency.
	
	\subsection{Exploiting the Low-Fisher Subspace for Lifelong Adaptation}
	The success of \textbf{OAL} beyond the static performance ceiling suggests that model capacity in edge-oriented perception systems should not be interpreted only in terms of total parameter count. The results indicate that even a compact \textbf{MobileNet}V3-Small retains a \textbf{low-Fisher} subspace that can support local online adaptation, while the \textbf{high-Fisher} parameters remain responsible for preserving global geometric structure.
	
\begin{figure*}[!htbp]
		\centering
		\includegraphics[width=0.9\linewidth]{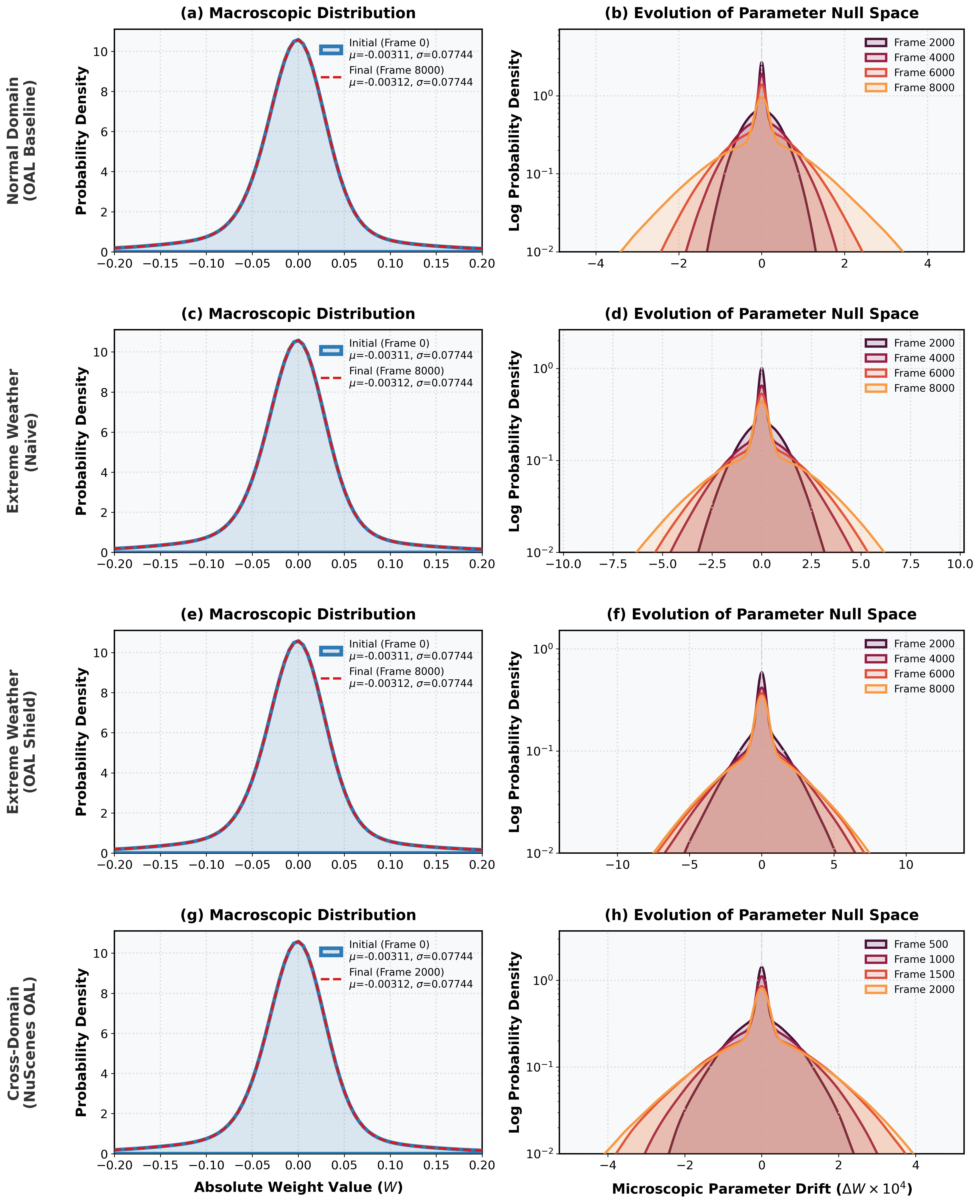}
		\caption{\textbf{Comprehensive parameter phase-space analysis across varying environments and learning strategies.} The left column (a, c, e, g) displays the macroscopic absolute weight distributions, demonstrating structural stability across all configurations. The right column (b, d, f, h) visualizes the corresponding microscopic parameter drift ($\Delta W$) within the null space. Notably, the unconstrained baseline (d) exhibits uniform parameter diffusion under extreme weather, characterized by a generalized Gaussian-like broadening. In contrast, the proposed OAL mechanism (f) enforces selective plasticity, characterized by a heavy-tailed drift distribution where critical parameters remain strictly concentrated at the center while redundant parameters expand to accommodate environmental variations. This regulated adaptation scales effectively to significant cross-domain shifts, such as the NuScenes dataset (h).}
		\label{fig:weight_evolution}
	\end{figure*}

	By anchoring the high-Fisher parameters, the proposed framework enforces \textit{selective plasticity} and redirects most online updates toward the more flexible low-Fisher subspace. The statistical evolution of this mechanism is visualized in Fig.~\ref{fig:weight_evolution}, where \textbf{EWC} limits broad parameter diffusion and concentrates large updates in the more redundant part of the parameter space, thereby improving stability under both severe environmental change and cross-domain shift.
	
	\subsection{Conclusion}
	This study shows that, for lightweight monocular depth estimation, improving deployment robustness cannot rely solely on stronger static offline optimization. The proposed \textbf{GCAF} mechanism improves structural efficiency under strict model-capacity constraints, while the \textbf{OAL} framework further enables stable continual adaptation through confidence-guided online learning and \textbf{EWC}-based selective plasticity.
	
	Future work will explore the integration of this continuous adaptation mechanism with pseudo-LiDAR generation pipelines, aiming to leverage the dynamically stabilized depth representations to enhance the accuracy and robustness of 3D object recognition in capacity-constrained autonomous systems.

	% ==========================================
	% ACKNOWLEDGMENT
	% ==========================================
	\section*{Acknowledgment}
	The authors declare no competing interests. Custom code, including framework implementations and pre-trained backbone weights required to reproduce all findings, will be made publicly available on GitHub upon acceptance. The datasets analyzed during this study are derived from public domain resources.

	% ==========================================
	% REFERENCES
	% ==========================================
	\bibliographystyle{IEEEtran}
	\bibliography{references} 
	
	\end{document}